\documentclass[twocolumn,10pt]{autart}  
\usepackage{graphicx}         
\usepackage{amsmath}
\usepackage{enumerate}      
\usepackage{tikz}
\usepackage{apacite}
\usepackage{natbib}
\usepackage{amsfonts}
\usepackage{doi}
\usepackage{amssymb}
\usepackage{subfigure}
\usepackage{hyperref}
\usepackage{algorithm}
\usepackage{algorithmic}
\hypersetup{
colorlinks = true,
citecolor = cyan,
}
\usepackage{verbatim}
\usepackage{wrapfig}

\newtheorem{remark}{Remark}

\newtheorem{assumption}{Assumption}
\newtheorem{theorem}{Theorem}

\begin{document}

\begin{frontmatter}

\title{Learning a Class of Mixed Linear Regressions: Global Convergence under General Data Conditions\thanksref{footnoteinfo}} 

\thanks[footnoteinfo]{This work was supported by Natural Science Foundation of China under Grants T2293772 and 12288201.}
\author[AMSS,CAS]{Yujing Liu}\ead{liuyujing@amss.ac.cn},     
\author[AMSS,CAS]{Zhixin Liu}\ead{lzx@amss.ac.cn},     
\author[AMSS,CAS]{Lei Guo}\ead{Lguo@amss.ac.cn}         

\address[AMSS]{Key Laboratory of Systems and Control, Academy of Mathematics and Systems Science,\\
Chinese Academy of Sciences, Beijing 100192, P. R. China.} 
\address[CAS]{School of Mathematical Sciences, University of Chinese Academy of Sciences, Beijing 100049, P. R. China.}          


\begin{abstract}  
Mixed linear regression (MLR) has attracted increasing attention because of its great theoretical and practical importance in capturing nonlinear relationships by utilizing a mixture of linear regression sub-models.
Although considerable efforts have been devoted to the learning problem of such systems, i.e., estimating data labels and identifying model parameters, most existing investigations employ the offline algorithm, impose the strict independent and identically distributed (i.i.d.) or persistent excitation (PE) conditions on the regressor data, and provide local convergence results only.
In this paper, we investigate the recursive estimation and data clustering problems for a class of stochastic MLRs with two components.
To address this inherently nonconvex optimization problem, we propose a novel two-step recursive identification algorithm to estimate the true parameters, where the direction vector and the scaling coefficient of the unknown parameters are estimated by the least squares and the expectation-maximization (EM) principles, respectively.
Under a general data condition, which is much weaker than the traditional i.i.d. and PE conditions, we establish the global convergence and the convergence rate of the proposed identification algorithm for the first time. 
Furthermore, we prove that, without any excitation condition on the regressor data, the data clustering performance including the cumulative mis-classification error and the within-cluster error can be optimal asymptotically.
Finally, we provide a numerical example to illustrate the performance of the proposed learning algorithm.
\end{abstract}

\end{frontmatter}

\section{Introduction}
Finite mixture models have attracted much attention in system identification and statistical analysis because they can provide convenient and efficient representations for modeling complex structure among the data  (\cite{mclach2019}).
Mixed linear regression (MLR), one of such models, is widely used for data classification tasks and has numerous applications in areas including trajectory clustering, health care analysis, face recognition and drug sensitivity prediction (cf., \cite{gaffney1999trajectory,zilber2023imbalanced,chai2007locally,li2019drug}).
Besides this, MLR can also be used to describe a wide range of prevalent systems in practice.
One such example is the piecewise affine systems in engineering (cf., \cite{bemporad2005bounded}), where the data labels depend on the system states.
Moreover, bilinear systems used in modeling of biological and economical systems can also be regarded as MLRs, provided that the input signal switches over a finite set, as in the case of bang-bang control (cf., \cite{pardalos2010optimization}).
Another noteworthy example is the fundamental switched linear models in the control community.
Specifically, existing investigations on the switched linear model can be divided into three categories:
i) The true parameter set is known, but the data label sequence is unknown (cf., \cite{xue2001necessary}); ii) The data label sequence is known, but the true parameter set is unknown (cf., \cite{cheng2005stabilization});
iii) Both the true parameter set and the data label sequence are unknown. 
Many important practical scenarios, such as learning of hydraulic pumping systems (\cite{Barbosa2019}), fall into this third category, which exactly corresponds to the MLR studied in this paper.

In the MLRs, each input-output observation data is generated by one of several stochastic linear regression submodels with unknown parameters, but we do not know which submodel the data comes from, meaning that the data label is unknown to us.
The main challenge in MLR is how to construct algorithms to identify the unknown submodel parameters and classify the newly observed data into its correct cluster, which is essentially an unsupervised learning problem (\cite{gan2007}).
Note that the coupling of parameter estimates and data label estimates makes the MLR learning problem particularly difficult. In fact, it has been proven to be NP-hard even without noise in the absence of any statistical assumptions on the data (\citet{yi2014alternating}). 

In order to investigate the learning problem of MLR with both unknown parameter set and unknown data label sequence, many estimation algorithms have been proposed from various perspectives:
\textit{i) Optimization-based methods}: Techniques such as mixed-integer linear programming (\cite{amaldi2016discrete}) and constrained rank minimization (\cite{ozay2015set}) have been used to the MLR learning problem.
\textit{ii) Tensor-based methods}: The parameter estimation problem has been reformulated as a problem of extracting a certain decomposition of a constructed symmetric tensor (cf., \cite{anandkumar2014tensor});
\textit{iii) Algebraic methods}: By transforming multiple subsystems into a single but more complex linear system, recursive estimation algorithms have been proposed to estimate the true parameters by differentiating the hybrid decoupling polynomial (cf., \cite{vidal2008recursive,bako2011recursive});
\textit{iv) Likelihood-based methods}: Various algorithms, including the classical population expectation-maximization (EM) algorithm (cf., \cite{,balakrishnan2017statistical}) and variational Bayesian methods incorporating available prior knowledge (\cite{ma2019parameter}), have been utilized to obtain the maximum likelihood estimator (MLE) of the unknown true parameters.

Based on the above algorithms, some theoretical results have been established for the MLR learning problem.
Here, we just name a few results based on the iterative algorithms, and more results can be found in survey papers (cf., \citet{ozay2015set,moradvandi2023models}).
For the MLR with two submodels where the unknown parameters are symmetric, local convergence results around the true parameters have been obtained by \citet{balakrishnan2017statistical} with the population EM algorithm to identify the true parameters and with the i.i.d. standard Gaussian assumption on the regressor.
Under the same algorithm and data assumption, a larger basin of attraction for local convergence has been provided in (\cite{klusowski2019estimating}) and subsequently, the global convergence results in a probabilistic sense have been established by
\citet{kwon2024global}.
For the MLR with multiple submodels and with no noise disturbance, the recursive estimation algorithms based on algebraic methods have been proposed, and then the global convergence results have been established under a stronger PE condition on the regressor data (cf., \cite{vidal2008recursive, bako2011recursive}). 
Moreover, for MLR disturbed by noises, based on the EM algorithm, local convergence results have been obtained under the assumption that the regressor is i.i.d. Gaussian (cf., \cite{balakrishnan2017statistical,kwon2020converges}) or stationary and ergodic with a PE condition (\cite{liu2024convergence}).

To summarize, the above mentioned theoretical investigations have the following limitations.
Firstly, most algorithms are of offline character, requiring all the input-output data at each iteration (cf., \citet{balakrishnan2017statistical}, \citet{kwon2024global}).
In comparison with offline algorithms, recursive algorithms are more desirable in practical applications because they can be updated conveniently based on the current estimate and new input-output data, without requiring storage of all previous data and with lower computational cost.
Although some recursive algorithms based on the algebraic methods have been proposed to identify the MLR (cf., \cite{vidal2008recursive}), these algorithms operate in a lifted space leading to a higher computational complexity.
Secondly, almost all existing studies require the PE condition or even i.i.d. Gaussian assumption on the data or regressor signals (cf., \cite{mu2022persistence,vidal2008recursive}).
These conditions are difficult to satisfy or verify in general, especially in stochastic uncertain systems with feedback control, where the input and output data are generally determined by nonlinear stochastic dynamic equations (\cite{lei2020feedback}).
Thirdly, the theoretical guarantees on the convergence of the MLR learning algorithms are mostly local (cf., \cite{balakrishnan2017statistical,klusowski2019estimating}) or global in the noiseless case (cf., \cite{vidal2008recursive}) with only a few exceptions (cf., \cite{kwon2024global,liu2023global}).
Furthermore, very few results are provided for the performance analysis of data clustering.
As far as we know, the investigation on the global convergence of the recursive identification algorithm  and data clustering for MLR under general data conditions is still lacking.

In this paper, we shall study the recursive identification and data clustering problems for a class of stochastic MLRs with two components and with noise disturbance. 
The main contributions of this paper are as follows:
\begin{itemize}
\item We propose a novel two-step recursive identification algorithm to estimate the unknown parameters of the stochastic MLR. The first step aims to estimate the direction vector by using the least squares (LS) method, and the second step tries to identify the scaling coefficient based on the EM principle by approximating the maximizer of the nonconvex likelihood function. 
\vskip 0.1in
\item By constructing a stochastic Lyapunov function and using the martingale estimation methods, we are able to, for the first time, establish the global convergence and convergence rate results for the MLR learning problem without the i.i.d. data assumptions or PE conditions, which makes our theory applicable to data generated by stochastic systems with feedback control.
\vskip 0.1in
\item Based on our proposed learning algorithm, we are able to show that the data clustering performance including the cumulative mis-classification error and within-in cluster error can be optimal asymptotically, without any excitation condition imposed on the data.
\end{itemize}

The remainder of this paper is organized as follows: In Section \ref{pf}, we present the problem formulation, including notations, model description, and the proposed learning algorithm. 
Section \ref{mainresults} states the main results on the global convergence of the learning algorithm and the data clustering performance. 
Sections \ref{proof} provide the proofs of the main results. 
Section \ref{simu} gives the numerical simulations to verify the effectiveness of our algorithm. 
Finally, we conclude the paper with some remarks in Section \ref{conclu}.

\section{Problem Formulation}\label{pf}
\subsection{Notations}
In this paper, $v\in{\mathbb{R}^d}$ denotes a $d$-dimensional column vector, $v^{\top}$ and $\|v\|$ are its transpose and Euclidean norm, respectively.
For a $d\times d$-dimensional symmetric matrix $A$, $\|A\|$ denotes the operator norm induced by the Euclidean norm, $\text{tr}(A)$ is the trace, $\lambda_{\max}(A)$ and $\lambda_{\min}(A)$ represent the maximum and minimum eigenvalues, respectively.
For two symmetric matrices $A$ and $B$, $A>(\geq)B$ indicates that $A-B$ is a positive (semi-positive)-definite matrix.
 Let $\{A_k\}$ be a sequence of matrices and $\{b_k\}$
be a positive scalar sequence. 
It is said that $A_k = O(b_k)$ if there exists a constant $C > 0$ such that $A_k \leq Cb_k$ holds for all $k \geq 0$,
and $A_k = o(b_k)$ means that $\lim\nolimits_{k\to\infty} \|A_k\|/b_k = 0$.

For a probability space $(\Omega, \mathcal{F},P)$, $\Omega$ is the sample space, $\mathcal{F}$ is the $\sigma$-algebra on $\Omega$ representing a collection of events, and $P$ is a probability measure defined on $(\Omega, \mathcal{F})$.
For an event $\mathcal{A}\in \mathcal{F}$, the complement of $\mathcal{A}$ is defined as $\mathcal{A}^c=\Omega-\mathcal{A}$.
The indicator function $\mathbb{I}_\mathcal{A}$ on $\Omega$ is defined as $\mathbb{I}_\mathcal{A}=1$ if the event $\mathcal{A}$ occurs, and $\mathbb{I}_\mathcal{A}=0$ otherwise.
An event $\mathcal{A}$ is said to occur almost surely (a.s.) if $P(\mathcal{A})=1$.
We use $\mathbb{E}\left[\cdot \right]$ to denote the mathematical expectation operator and $\mathbb{E}[\cdot|\mathcal{F}]$ to represent the conditional expectation operator given the $\sigma$-algebra $\mathcal{F}$.  
Let $\{\mathcal{F}_k, k\geq0\}$ be a non-decreasing sequence of $\sigma$-algebras.
A sequence of random variables $\{x_{k}\}$ is said  to be adapted to $\{\mathcal{F}_k\}$ if $x_k$ is $\mathcal{F}_k$-measurable for all $k\geq 0$. Furthermore, if $\mathbb{E}[x_{k+1}|\mathcal{F}_k]=0$ for all $k\geq0$, then the adapted sequence $\{x_{k}, \mathcal{F}_k\}$ is called a martingale difference sequence.

\subsection{Model Description}
This paper considers the following MLR with two submodels:
\begin{equation}\label{modelx}
y_{n+1}=z_n\beta^{*\top}\phi_n+w_{n+1},\\
\end{equation}
where $\beta^{*}\in\mathbb{R}^{d}$ is an unknown system parameter to be estimated, $y_n\in\mathbb{R}$, $\phi_{n}\in\mathbb{R}^d$ and $w_{n}\in\mathbb{R}$ are the output, regressor vector and noise disturbance at time $n$, respectively.  
Besides, the data label $z_n$ is a hidden variable indicating which submodel the data $\{\phi_n,y_{n+1}\}$ belongs to at time $n$. 
We note that the fundamental phase retrieval problem in physics (\cite{shechtman2015phase}) and the absolute activation function learning problem in machine learning (\cite{chen2019gradient}) are both related to the MLR (\ref{modelx}), which has been widely studied in the literature (cf., \cite{balakrishnan2017statistical,klusowski2019estimating,kwon2024global}).

In this paper, we  will first propose a recursive identification algorithm to estimate the true parameter $\beta^*$ based on the streaming input-output data $\{\phi_n,y_{n+1}\}_{n=1}^{\infty}$, and establish convergence results for the proposed identification algorithm with a general data condition.
Then, based on the above estimate of $\beta^*$, we will investigate the performance that the newly emerged data can be classified into the correct submodel.

Before presenting the assumptions used in this paper, we introduce the following two sequences of $\sigma$-algebras:
\begin{equation*}
\begin{aligned}
& \mathcal{F}_n=\sigma\{\phi_{k},w_{k},z_k,k\leq n\},\\ &  \mathcal{F}'_n=\sigma\{\phi_{k},w_{k},z_{k-1},k\leq n\}.
\end{aligned}
\end{equation*}

\begin{assumption}\label{asm2}
Given $\mathcal{F}'_n$, the hidden variable $z_n$ follows the conditional distribution $P(z_n=1)=p$ and $P(z_n=-1)=1-p$ with $p$ being an unknown constant in $[0,1]\backslash\{\frac{1}{2}\}$. 
\end{assumption}
 
\begin{remark}
It is clear that the parameter $p=\frac{1}{2}$ for the MLR has a Lebesgue measure zero over $[0,1]$. In practice, real-world data are likely unbalanced (i.e., $p\ne \frac{1}{2}$), as observed in the medicine insurance cost prediction and the outlier detection problems (\cite{hawkins1984location,zilber2023imbalanced}).
However, the balanced case $p=\frac{1}{2}$ is also of significance, since it corresponds to the maximum ``uncertainty'' in the hidden variables. 
Without the i.i.d. data assumption, the balanced case can also be investigated (see, \cite{liu2023global}), but needs the data sequence to be stationary and ergodic to guarantee convergence. 
\end{remark}

\begin{assumption}\label{asm3}
Given $\mathcal{F}_n$, the noise $w_{n+1}$ follows a conditional Gaussian distribution $\mathcal{N}(0,\sigma^2)$.
\end{assumption}
\begin{assumption}\label{asm4}
There exists a constant $\delta\in[0,\frac{1}{2})$ such that as $n\to\infty$,
\begin{equation}
\|\phi_{n}\|^2=O(n^{\delta}), \ \hbox{a.s.}
\end{equation}
\end{assumption}
\begin{remark}
We provide several commonly-used examples (\cite{chen2}) that satisfy Assumption \ref{asm4}: 

1) If $\{\phi_n\}$ is bounded, then the parameter $\delta$ can be zero;

2) If $\{\phi_n\}$ follows a Gaussian distribution as used in the previous investigations on MLR learning problems, then $\|\phi_n\|^2=O(\log n)$ and the parameter $\delta$ can be chosen as any constant in $(0,\frac{1}{2})$; 

3) If there is a constant $\beta>4$ such that $\sup\limits_{n\geq0}\mathbb{E}[\|\phi_n\|^{\beta}]<\infty,$ 
then the parameter $\delta$ can be any constant in $\left(\frac{2}{\beta},\frac{1}{2}\right)$.
\end{remark}

\subsection{Recursive Identification Algorithm}
In this subsection, we design a two-step recursive identification algorithm to estimate the true parameter $\beta^*$.

First, by model (\ref{modelx}) and Assumptions \ref{asm2}-\ref{asm3}, we can obtain
\begin{equation}\label{yk}
\mathbb{E}\left[y_{n+1}|\mathcal{F}'_n\right]=\mathbb{E}\left[z_{n}|\mathcal{F}'_n\right]\beta^{*\top}\phi_n=\theta^{*\top}\phi_n,
\end{equation}
where \begin{equation}\label{theta*}
\theta^*\triangleq(2p-1)\beta^*.
\end{equation} 
It is evident that (\ref{yk}) is actually a linear regression model and the parameter $\theta^*$ can be estimated by the classical recursive LS algorithm under general data conditions (see, e.g., \cite{guo1995convergence}). 

Second, we define a scaling coefficient between the norms of $\theta^*$ and $\beta^*$ as
\begin{equation}
q^*=\frac{\|\beta^*\|}{\|\theta^*\|}.
\end{equation}
Based on the principle of EM as used in (cf., \cite{liu2023global,kwon2024global}), one may initially consider the estimate of the scaling parameter $q^*$ as follows:
\begin{equation}\label{offline}
\begin{aligned}
&q_{n+1}\\
=&\left[\sum\limits_{k=1}^n (\theta_k^{\top}\phi_k)^2\right]^{-1}\left[\sum\limits_{k=1}^ny_{k+1}\tanh\big(\frac{q_k \theta_k^{\top}\phi_ky_{k+1}}{\sigma^2}\big)\right].
\end{aligned}
\end{equation}
Following the derivation of the recursive LS, it is easy to obtain the equivalent expression of (\ref{offline}) in an online version.
Due to the possible unboundedness of the regressor vectors and the parameter estimates, we need to make some modifications to (\ref{offline}) in order to carry out a rigorous theoretical analysis.

First, inspired by the idea of weighted LS algorithm (\cite{guo1996self}), we introduce a weighted scalar sequence $\{n^{\delta}\}$ in the adaptation gain to deal with the difficulties caused by the possible unboundedness of the regressor $\{\phi_n\}$. 

Second, we design the adaptation gain $\{\alpha_n\}$ in estimating the scaling coefficient by using the lower bound of the derivative of the innovation term in (\ref{offline}), which facilitates the convergence of parameter estimates.

Third, we introduce the following projection operator to constrain the growth rate of the estimates $q_n$ and to constrain the distance of $q_n$ to the local minimizer $0$:
\begin{equation}\label{pid}
\Pi_{D_n}(x)=\mathop{\arg\min}_{y\in D_n} |x-y|, \forall x\in \mathbb{R},
\end{equation}
where
$D_n=\{x:1 \leq x\leq \sqrt{\log (n+e)} \}$ with $1$ being the lower bound of $q^*$ by Assumption \ref{asm3}.

Based on the above analysis, our recursive identification algorithm can be summarized as Algorithm \ref{alg1} below. 
\begin{algorithm}
\caption{Two-step recursive identification algorithm}\label{alg1}
\vskip -0.1in
\begin{subequations}
\begin{align}
&\text{\textbf{Input: }The initial values}\ \theta_0\ne 0, \beta_0\ne0 \ \text{and}\ P_0>0 \nonumber \\
&\ \ \text{can be chosen arbitrarily, and}\ r_0=1.\nonumber\\
&\textbf{Step 1:}  \text{ Estimate of the vector $\theta^*$}\nonumber\\
&\ \ \theta_{n+1}=\theta_n+a_nP_n\phi_n\left(y_{n+1}-{\theta}_{n}^{\top}{\phi}_{n}\right),\label{theta1}\\
&\ \ P_{n+1}=P_n-a_n P_n\phi_n\phi_n^{\top}P_n,\label{theta2}\\
&\ \ a_n=\frac{1}{n^{\delta}+\phi_n^{\top}P_n\phi_n},\label{theta3}\\
&\textbf{Step 2:}\  \text{Estimate of the scaling coefficient $q^*$}\nonumber\\
&\ \ q_{n+1}=\Pi_{D_n}\left[q_{n}+\frac{\alpha_{n}\theta_n^{\top}\phi_{n}}{n^{\delta} r_{n+1}}\left(s_{n+1}-q_{n}\theta_n^{\top}\phi_{n}\right)\right],\label{beta}\\
&\ \ s_{n+1}=y_{n+1}\tanh\left(\frac{q_n\theta_n^{\top}\phi_ny_{n+1}}{\sigma^2}\right),\\
&\ \ r_{n+1}=r_{n}+\frac{\alpha_{n}^2}{n^{\delta}}\left(\theta_n^{\top}\phi_{n}\right)^2,\label{rk}\\
&\ \ \alpha_{n}=1-\exp\left(-\frac{(\theta_n^{\top}\phi_n)^2}{2\sigma^2}\right)\label{alphak},\\
&\textbf{Output:}  \text{ Estimate of}\ \beta^*\text{sgn}(2p-1) \nonumber\\
&\ \ \beta_{n+1}=q_{n+1}\theta_{n+1}.
\end{align}
\end{subequations}
\vskip -0.1in
\end{algorithm}

\section{Main Results}\label{mainresults}
In this section, we give some theoretical results of the recursive identification algorithm (i.e., Algorithm \ref{alg1}). To be specific, we will first establish the global convergence and convergence rate of the parameter estimate $\beta_n$, and then provide the performance analysis for data clustering based on the proposed identification algorithm.	

Denote the estimation error of the algorithm as 
$$\tilde{\beta}_n=\beta^{*}\text{sgn}(2p-1)-\beta_n,$$
where $\beta^{*}\text{sgn}(2p-1)$ is the true parameter of the submodel with larger mixture probability.
The following theorem provides an upper bound of the estimation error under a non-PE data condition.
\begin{theorem}\label{theorem1}
Under Assumptions \ref{asm2}-\ref{asm4}, the estimation error generated by Algorithm \ref{alg1} has the following upper bound as $n\to\infty$:
\begin{equation}\label{equ}        \|\tilde{\beta}_{n+1}\|^2=O\left(\frac{n^{\kappa}}{\lambda_{\min}(n)}\right), \ \hbox{a.s.,}
\end{equation}
where $\kappa=\max\left\{\frac{1}{2}+\delta+\varepsilon,\frac{2+\delta}{3}+\varepsilon\right\}<1$ with $\delta\in [0, \frac{1}{2})$ being defined in Assumption \ref{asm4} and $\varepsilon$ being an arbitrarily small positive constant, and 
\begin{equation}\label{lamin}
\lambda_{\min}(n)=\lambda_{\min}\left(P_0^{-1}+\sum\limits_{k=1}^n\phi_k\phi_k^{\top}\right).
\end{equation}
\end{theorem}
The detailed proof of Theorem \ref{theorem1} is provided in Section \ref{proof}.
\begin{remark}
From Theorem \ref{theorem1}, we know that if 
\begin{equation}\label{condition}
n^{\kappa}=o(\lambda_{\min}(n)), \ \hbox{a.s.,}
\end{equation}
then the estimate $\beta_n$ will converge to the true parameter $\beta^*\text{sgn}(2p-1)$ almost surely.
The excitation condition (\ref{condition}) on the regressor data is significantly weaker than the i.i.d. Gaussian assumption used in most existing investigations on the MLR learning problem (cf., \cite{kwon2024global,zilber2023imbalanced}) and even weaker than the following traditional PE condition: 
$$
n=O(\lambda_{\min}(n)), \ \hbox{a.s.}
$$
\end{remark}
For the new data $\{\phi_n,y_{n+1}\}$, if $z_n=1$, we denote the index of the correct cluster as $\mathcal{I}_n=1$, and if $z_n=-1$, we denote it as $\mathcal{I}_n=2$ .
If the true parameter $\beta^*$ is known, then the optimal estimate for $\mathcal{I}_n$ is as follows:
\begin{equation}
\mathcal{I}_n^{*}=\mathop{\arg\min}_{i=1,2}\{(y_{n+1}+(-1)^{i}\beta^{*\top}\phi_n)^2\}.
\end{equation}
By replacing the true parameter $\beta^*$ by its estimate $\beta_n$ generated by Algorithm \ref{alg1}, we can obtain the estimate for $\mathcal{I}_n$:
\begin{equation}\label{class}
\hat{\mathcal{I}_n}=\mathop{\arg\min}_{i=1,2}\{(y_{n+1}+(-1)^{i}\beta_{n}^{\top}\phi_n)^2\}.
\end{equation}
We now provide an upper bound of the cumulative mis-classification error in the following theorem:
\begin{theorem}\label{theorem2}
Let Assumptions \ref{asm2}-\ref{asm4} be satisfied. Then the average of the cumulative mis-classification error has the following upper bound as $n\to\infty$:
\begin{equation}\label{them4.21}
\bigg|\frac{1}{n}\sum\limits_{k=1}^n \mathbb{I}_{\{\hat{\mathcal{I}_k}\ne\mathcal{I}_k\}}-\frac{1}{n}\sum\limits_{k=1}^n\mathbb{I}_{\{\mathcal{I}_k^*\ne\mathcal{I}_k\}}\bigg|=O\left(\sqrt{\frac{\log n}{n^{1-\delta}}}\right),\ \hbox{a.s.,}
\end{equation}
where $\delta\in [0, \frac{1}{2})$ is defined in Assumption \ref{asm4}.
\end{theorem}
To further evaluate the data clustering performance, we also introduce a commonly-used within-cluster error index (cf., \cite{zilber2023imbalanced}) as follows:
\begin{equation}\label{wce}
J_n=\sum\limits_{k=1}^n(y_{k+1}+(-1)^{\hat{\mathcal{I}_k}}\beta_{k}^{\top}\phi_k)^2,
\end{equation}
whose upper bound is provided in the following theorem:
\begin{theorem}\label{theorem3}
Let Assumptions \ref{asm2}-\ref{asm4} be satisfied. Then the within-cluster error has the following upper bound as $n\to\infty$:
\begin{equation}\label{them4.22}
\frac{J_n}{n} \leq \sigma^2+O\left(n^{-\frac{1-\kappa}{2}}\right),  \ \hbox{a.s.,}
\end{equation}
where $\kappa<1$ is defined in Theorem \ref{theorem1}.
\end{theorem}

The proofs of Theorems \ref{theorem2}-\ref{theorem3} are provided in Section \ref{proof}.
\begin{remark}
In Theorems \ref{theorem2}-\ref{theorem3}, without any excitation condition on the data $\{\phi_n\}$, we demonstrate that the data cluster performance including the cumulative mis-classification error and the within-cluster error can be optimal asymptotically.
This is reminiscent of the result for the classical LS algorithm for linear regression models, where the adaptive predictors can attain the optimal performance without requiring any excitation condition on the system data (cf., \cite{guo1995convergence}).
\end{remark}

\section{Numerical Simulation}\label{simu}
Consider the following stochastic dynamical system:
\begin{equation}\label{model4.5}
\begin{aligned}
&y_{n+1}=z_n\beta^{*\top}\phi_n+w_{n+1},\\
&\phi_{n+1}=0.8\phi_n+\frac{1}{n^{1/10}}e_{n+1},
\end{aligned}
\end{equation}
where the hidden variable sequence $\{z_n\}$ is i.i.d. with $P(z_n=1)=0.6$ and $P(z_n=-1)=0.4$, the noise sequence $\{w_{n+1}\}$ is i.i.d. with $\mathcal{N}(0,1)$ and $\{e_{n+1}\}$ is i.i.d. with $\mathcal{N}(0,I)$. Besides, $\{w_{n+1}\}$, $\{e_{n+1}\}$ and $\{z_n\}$ mutually independent.

It is obvious that the regressor $\phi_n$ is not i.i.d. and as $n\to\infty$, $\{\phi_n\}$ satisfies the following properties:
\begin{equation}\label{simu4.1}
\begin{aligned}
&n^{4/5}=O\left(\lambda_{\min}\left(\sum\limits_{k=1}^n\phi_k\phi_k^{\top}\right)\right), \\
&\lambda_{\max}\left(\sum\limits_{k=1}^n\phi_k\phi_k^{\top}\right)=O(n), \ \hbox{a.s.,}
\end{aligned}
\end{equation}
which demonstrates that $\{\phi_n\}$ does not satisfy the PE condition used in most investigations on the MLR learning problem. 

\begin{figure}
\centering
\includegraphics[width=0.475\textwidth]{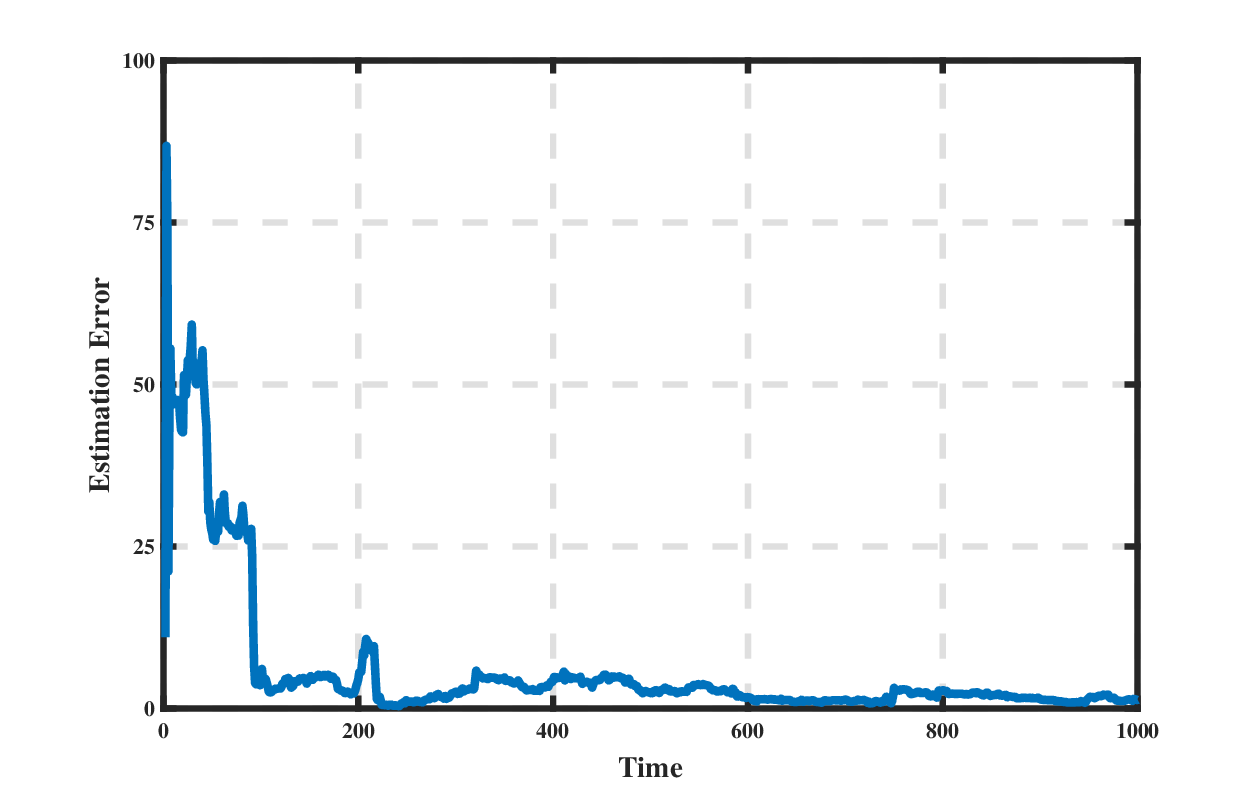}
\caption{Trajectory of the estimation error $\|\tilde\beta_n\|^2$}
\label{fig4.1a}
\end{figure}

From Fig.\ref{fig4.1a}, we see that  the estimation error tends to 0 at $n$ tends to infinity,  which demonstrates the convergence of Algorithm \ref{alg1}.

\begin{figure}
\includegraphics[width=0.47\textwidth]{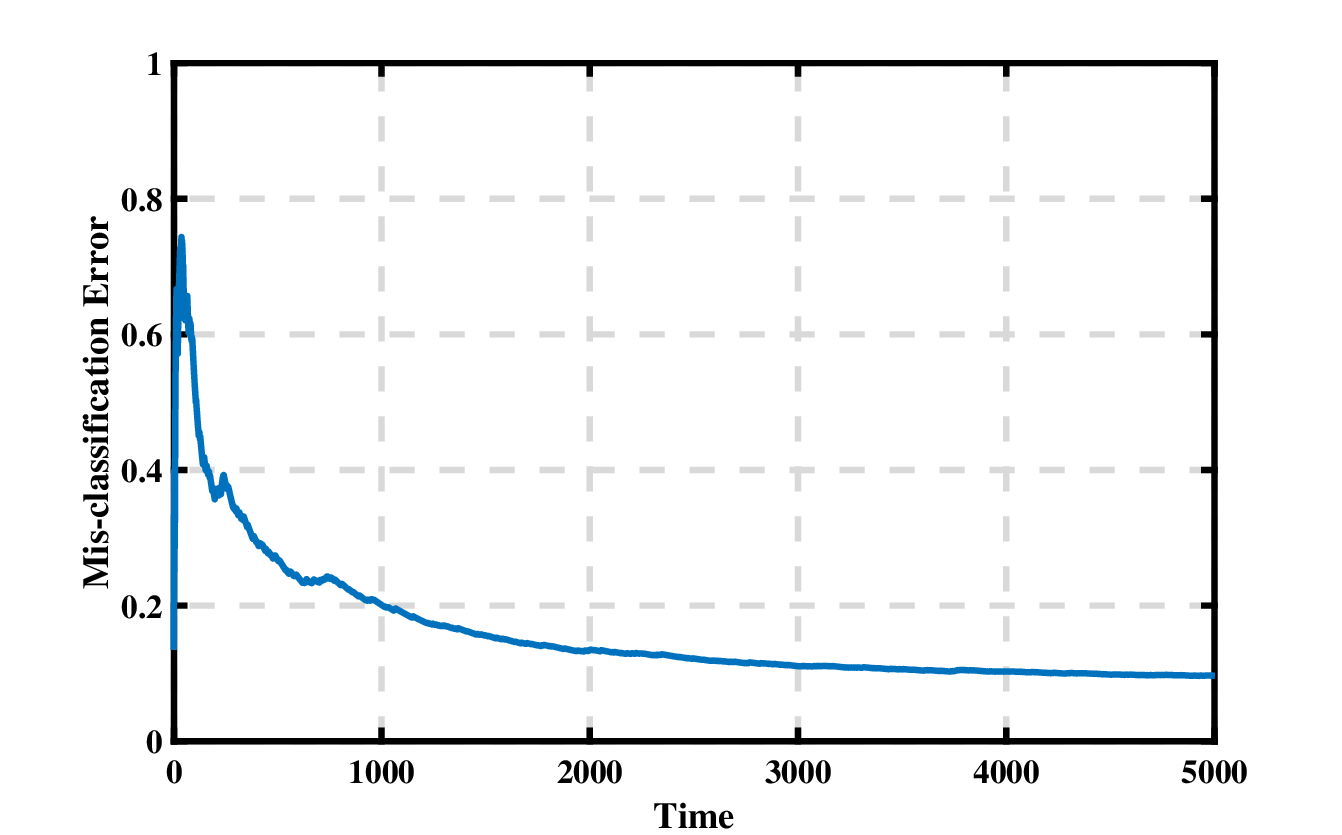}
\caption{Trajectory of $\frac{1}{n}\bigg|\sum\limits_{k=1}^n \mathbb{I}_{\{\hat{\mathcal{I}_k}\ne\mathcal{I}_k\}}-\sum\limits_{k=1}^n\mathbb{I}_{\{\mathcal{I}_k^*\ne\mathcal{I}_k\}}\bigg|$}
\label{fig4.3}
\end{figure}

\begin{figure}
\includegraphics[width=0.475\textwidth]{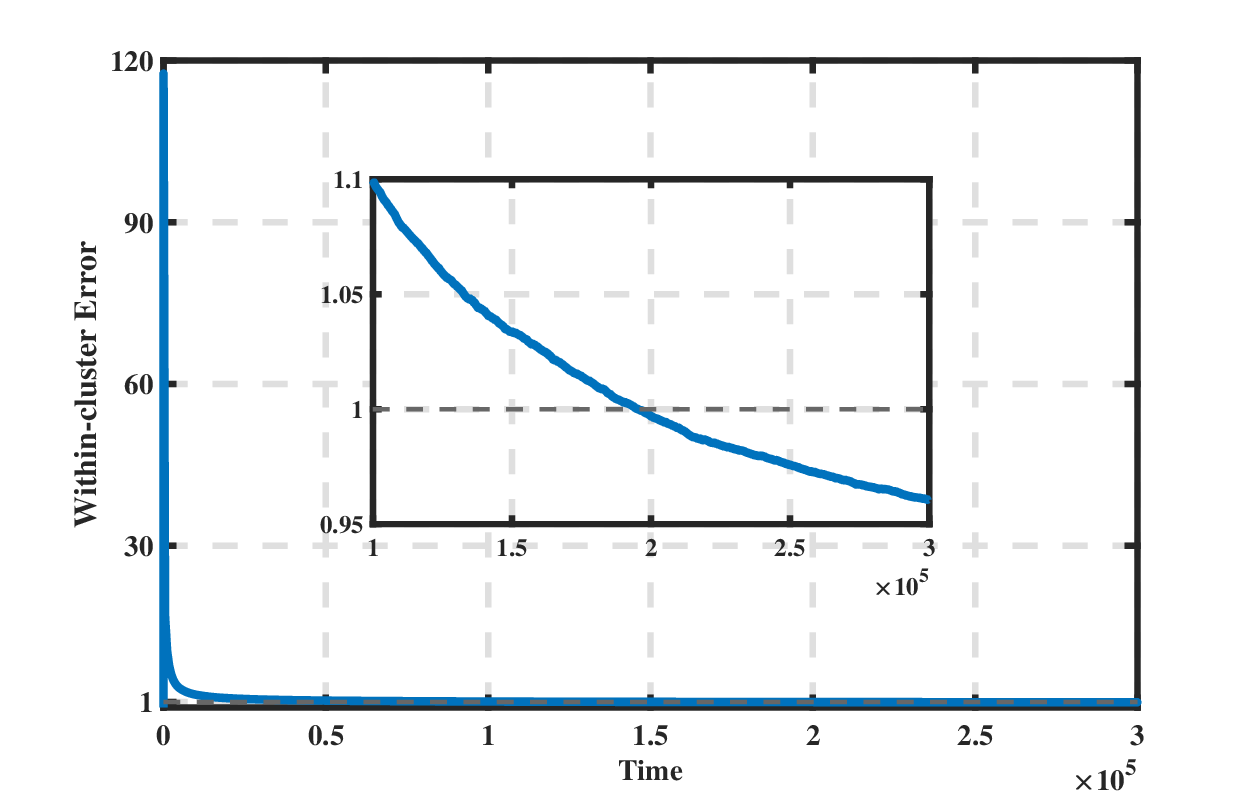}
\caption{Trajectory of $\frac{J_n}{n}$}
\label{fig4.1b}
\end{figure}

We further studied  the data clustering performance which are shown in Fig.\ref{fig4.3} and Fig.\ref{fig4.1b}.  From these two figures, we see that  the average of the cumulative mis-classification error will tend to zero as $n$ tends to $\infty$, and the average of the within-cluster error will decay 
below the noise variance $\sigma^2=1$ for large $n$.

\section{Concluding Remarks}\label{conclu}
This paper has proposed a novel two-step recursive identification algorithm for an unbalanced  symmetric MLR learning problem.
The proposed algorithm transforms the original nonconvex maximization problem for a high-dimensional likelihood function into a combination of an identification problem for a linear regression model with a nonconvex maximization problem for a one-dimensional likelihood function, thereby significantly weakening the difficulty in establishing the convergence of the parameter estimates.
Under a considerably general non-PE condition on the data that may come from feedback control systems, we have established the global convergence and convergence rate of the recursive identification algorithm.
Furthermore, we have showed that, without any excitation condition imposed on the data, the data clustering performance can be optimal asymptotically.
For further investigations, many interesting problems still need to be resolved in theory, e.g., how to establish a global convergence theory for learning problems of more complicated MLRs under general data conditions.

\bibliographystyle{apalike}        %
\bibliography{auto}
\end{document}